# YouTube for Patient Education: A Deep Learning Approach for Understanding Medical Knowledge from User-Generated Videos

Extended Abstract

**Xiao Liu**
University of Utah
Salt Lake City, UT, USA
xiao.liu@eccles.utah.edu

**Bin Zhang**
University of Arizona
Tucson, AZ, USA
binzhang@email.arizona.edu

**Anjana Susarla**
Michigan State University
Lansing, MI, USA
asusarla@broad.msu.edu

**Rema Padman**
Carnegie Mellon University
Pittsburgh, PA, USA
rpadman@cmu.edu

## ABSTRACT

YouTube presents an unprecedented opportunity to explore how machine learning methods can improve healthcare information dissemination. We propose an interdisciplinary lens that synthesizes machine learning methods with healthcare informatics themes to address the critical issue of developing a scalable algorithmic solution to evaluate videos from a health literacy and patient education perspective. We develop a deep learning method to understand the level of medical knowledge encoded in YouTube videos. Preliminary results suggest that we can extract medical knowledge from YouTube videos and classify videos according to the embedded knowledge with satisfying performance. Deep learning methods show great promise in knowledge extraction, natural language understanding, and image classification, especially in an era of patient-centric care and precision medicine.

## KEYWORDS

Health literacy, patient education, user-generated content, YouTube, deep learning.

## 1 INTRODUCTION

Health literacy has been defined as the degree to which individuals have the "capacity to obtain, process and understand basic health information and services needed to make appropriate health decisions. "[1] Estimates indicate that only 12% of the US population has proficient health literacy, and most adults have difficulty interpreting and using health information, which translates to poorer health outcomes and higher healthcare utilization and costs.

According to surveys by the Pew Research Center in 2006, 2010 and 2018, respectively, patients increasingly rely on the Internet to educate themselves through online health information searches. However, purely text-based medical instructions result in inadequate patient attention, comprehension, recall, and adherence, especially for patients with low literacy levels. YouTube hosts videos containing information on the pathogenesis, diagnosis, treatments, and prevention of various conditions. These health-related videos may offer a practical approach to bridge the health literacy gap [1]. However, healthcare videos on YouTube range from homemade ones expressing personal opinions and experiences to those made by reputable healthcare entities and expert clinicians for professional education [2]. While visual representations expand human cognition, several healthcare providers and government agencies have expressed concerns about the quality and reliability of healthcare information available through YouTube [2]. Hence, such sites could also misinform patients through incorrect or imprecise medical knowledge.

YouTube presents an unprecedented opportunity to explore how machine-learning methods can improve healthcare information dissemination. We propose an interdisciplinary lens that synthesizes machine learning methods with healthcare informatics themes to address the critical issue of developing a scalable algorithmic solution to evaluate videos from a health literacy and patient education perspective. In this preliminary study, we develop a deep learning method to understand the level of medical knowledge encoded in YouTube videos. We conduct our analysis on videos related to Diabetes, which is among the most prevalent and manageable of chronic diseases.

## 2 RESEARCH METHOD

We develop an analytical approach as depicted in Fig. 1 to understand the medical knowledge encoded in YouTube videos.

---
[1] https://health.gov/communication/literacy/issuebrief

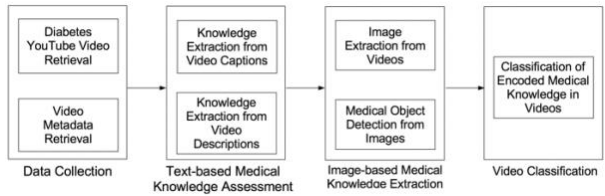

**Figure 1. Study Design for Understanding the Medical Knowledge in YouTube Videos**

## 2.1 Data Collection

We fetch the top 100 videos for each search term and store the ranking of returned videos and their metadata in a database for further analysis. In total, we collected 19,873 unique videos using over 200 search terms. The videos are contributed by both individual users and reputable healthcare organizations such as the Mayo Clinic, the American Diabetes Association, and the American Nutrition Association. We download a subset of 600 videos, which contain closed captions submitted by content contributors. This retrieved video dataset is used for video frame analysis and text analytics with video captions in robustness checks. Two graduate research associates reviewed these 600 videos, their descriptions, and captions. They provide the annotation for medical terminologies in video descriptions and captions and label the videos as high or low medical knowledge. A domain expert (a medical doctor) viewed all the annotations and consolidated the annotation results. The inter-rater reliability for annotating medical terms is 0.87. The inter-rater reliability for classifying encoded medical knowledge in videos is 0.92.

## 2.2 Text-based Medical Knowledge Extraction

We develop our text-based medical knowledge extraction approach based on a vector representation of words with word embedding and Bidirectional Long Short-Term Memory (BLSTM) Recurrent Neural Network to extract medical knowledge from the video descriptions and assess the level of medical knowledge. The following Algorithm 1 describes the proposed medical term extraction process.

| Algorithm 1. Medical Term Extraction from YouTube Videos |
|---|
| **Input**: A collection of sentences $C$.<br>A sentence $X \in C$ is represented as $[x_1, x_2, ..., x_n]$. $x_i$ refers to the $i^{th}$ word in the sentence $X$.<br>**Output**: A collection of labeled sentences.<br>Each sentence $X \in C$ is labeled as $L = [l_1, l_2, ..., l_n]$. $l_i$ refers to the label for the $i^{th}$ word in the sentence. $l_i \in \{NA, MT\}$. MT refers to a medical term. NA refers to a non-medical term.<br>**Procedures:**<br>1. Train a word-embedding model $W$ with all $X \in C$.<br>2. For each $w_t \in V$, $V$ is the vocabulary (unique words) in collection $C$, obtain the word embedding $W_t$ of $w_t$.<br>3. Create an annotated data set $C' \subset C$. $C'$ contains a collection of sentences $S$ and their labels $L'$. Each $w$ in $C'$ is represented by its embedding $W$.<br>4. Build a Bidirectional Long Short-Term Memory RNN model $L = R(S)$ with a collection of annotated dataset $C'$.<br>5. For each $X \in \{C - C'\}$ (unlabeled sentences), apply the model $R$ to obtain the labels for the sentence with $L' = R(X)$. |

To identify medical terms, we train an embedding model using the Skip-gram method in Word2vec [3] on the entire collection of video descriptions. To avoid rare words negatively affecting the model performance, we prune the vocabulary by replacing the less frequent words with a unified symbol "UNK" (short for the unknown token). We keep the top 5,000 frequent words in their original form and replace the remaining words with UNK. Identifying medical terminologies in video descriptions can be considered as a named entity recognition task. We devise a Bidirectional Long Short-Term Memory (BLSTM) model [4] to extract medical terms from the user-generated video descriptions at the sentence level. The video descriptions are first tokenized into individual sentences and words. Fig. 2 illustrates the BLSTM architecture for medical knowledge extraction.

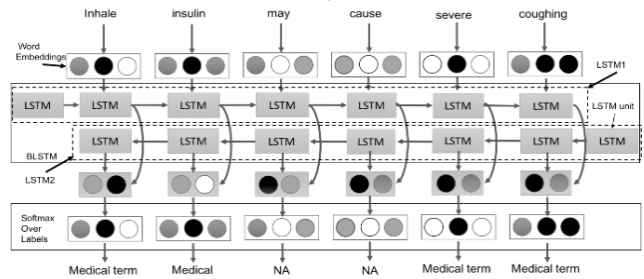

**Figure 2. BLSTM Architecture for Medical Knowledge Extraction from Text**

We train a 50-dimensional word-embedding model, meaning each word is converted to a 50-dimensional semantic vector. Instead of using a large hidden layer, we use 150 neurons in the BLSTM layer to avoid over-fitting. The outputs of the BLSTM layers are then processed by a Softmax classifier, which predicts the semantic type of each word in the input sentence. We apply the BLSTM RNN model to the closed captions from 600 annotated videos to extract medical terms from the captions.

## 2.3 Image-based Medical Knowledge Extraction

The performance of object detection in images has been significantly improved with the advent of the large datasets such as ImageNet using Convolutional Neural Network (CNN) architectures such as AlexNet, Inception, and ResNet. We examine if CNN can yield good performance on object detection in video frames and extract medical knowledge from the videos. We construct a large video frame dataset by sampling one frame every two seconds from the 600 annotated video dataset. In total there are 204,713 frames extracted. Our task is to predict the objects in each frame. We utilize an Inception CNN architecture that was pre-trained on approximately 1.28 million images (1,000 object categories) from the 2014 ImageNet Large Scale Visual Recognition Challenge [5]. The image classification is performed on a workstation with 32G RAM and NVIDIA Quadro K420 GPU. A batch of 100 frames requires 320 seconds to process. We predict the probability of the 1000 object categories for each frame using the Inception CNN architecture. We extract the top 5 object categories for each frame. Object category labels with low probability are not considered for further analysis. We identify the number of medically relevant objects at the video level.

## 2.4 Classification of Encoded Medical Knowledge in Videos

We train a classifier with content related video level features to classify videos into high and low medical knowledge videos. We apply the BLSTM RNN model to the closed captions from 600 annotated videos to extract medical terms from the captions. We incorporate the number of medically relevant objects as an additional feature in the logistic regression for medical knowledge classification. Among these annotated videos, 377 of them are labeled as high medical knowledge and 223 are labeled as low medical knowledge. We construct the logistic regression model for medical knowledge classification with the medical terms from closed captions video descriptions and medical objects from video frames. The model is trained on 480 videos and evaluated on 120 videos.

## 3 PRELIMINARY RESULTS

In text-based knowledge extraction, we compare our proposed method, which has achieved 92. 9% in f-measure, with three state-of-the-art baseline methods on the same task– a lexicon-based medical term extraction with MetaMap (28.9% in f-measure), a statistical Conditional Random Field model (81.6% in f-measure), and a standard recurrent neural network model (89.7% in f-measure). The results demonstrate the superior performance of our proposed approach. In image-based knowledge extraction, we extract the top 5 object categories for each frame. Object category labels with low probability are not considered for further analysis. We identify the number of medically relevant objects at the video level. The overall accuracy of the logistic regression model for medical knowledge classification is 85%. Detailed performance evaluation results on precision, recall, and f-measure are reported in Table 1 below.

**Table 1. Medical Knowledge Classification Evaluation Results**

|  | Precision | Recall | F-measure |
|---|---|---|---|
| High Medical Knowledge Videos | 86.5% | 80.5% | 83.4% |
| Low Medical Knowledge Videos | 83.8% | 84.3% | 84.0% |

## 4 POTENTIAL CONTRIBUTIONS AND FUTURE DIRECTIONS

This study is among the first to develop a deep learning approach to extract complex medical knowledge from real-world, user-generated video data. We approach the knowledge extraction and classification of videos from two different aspects: knowledge extraction from text (i.e., video descriptions and closed captions) and knowledge extraction from video frames. Knowledge extraction from video requires an understanding of the semantic meaning of the context, which poses more challenges than the current state of the art applications for medical image classification. Our current image analysis utilizes a pre-trained Inception model with the ImageNet dataset. The results are not tailored to the medical context. To extend our current work, we could focus on developing an integrated learning framework for video understanding with three different sources of data: video metadata, captions, and frames. More sophisticated deep learning models could improve the performance of automatic transcription generation and medical object detection.

In this preliminary study, we only assess the videos according to two categories: high and low medical knowledge.  In future work, we will apply more effort in categorizing the videos into several levels to address the needs of a diverse population, and defined more precisely by drawing on the vast literature on health literacy.

A potential contribution is in combining machine-learning methods with themes emphasized in the literature on human-computer interaction (HCI) to design recommender systems to retrieve medically relevant user-generated content better. Our study attempts to classify patient educational videos, increasing the ease with which patients can find helpful health information on YouTube. Our approach can flag misinformation through videos (i.e., videos with shallow medical knowledge) in the healthcare field and potentially apply to other domain as well. We can potentially improve both the process of video search as well as the process by which content can be identified or designed to enhance health literacy.